\documentclass[runningheads]{llncs}
\usepackage{graphicx}

\usepackage[ruled,vlined,linesnumbered]{algorithm2e}
\usepackage{float}
\usepackage{tabularx}
\usepackage{multicol}

\usepackage{enumitem}
\usepackage{comment}

\usepackage{hyperref}
\usepackage{subfigure}
\usepackage{amsmath}
\usepackage{calc}
\usepackage{wrapfig}
\usepackage{amssymb}
\usepackage{enumitem}

\begin{document}
\title{Sparsification for Fast Optimal Multi-Robot Path Planning in Lazy Compilation Schemes}
\titlerunning{Sparsification for Fast Optimal Multi-Robot Path Planning ...}
\author{Pavel Surynek\orcidID{0000-0001-7200-0542}}
%
\authorrunning{P. Surynek}

%
\institute{
Faculty of Information Technology\\
Czech Technical University in Prague\\
Th\'{a}kurova 9, 160 00 Praha 6, Czechia\\
\email{pavel.surynek@fit.cvut.cz}
}
\maketitle              
\begin{abstract}
Path planning for multiple robots (MRPP) represents a task of finding non-colliding paths for robots through which they can navigate from their initial positions to specified goal positions. The problem is usually modeled using undirected graphs where robots move between vertices across edges. Contemporary optimal solving algorithms include dedicated search-based methods, that solve the problem directly, and compilation-based algorithms that reduce MRPP to a different formalism for which an efficient solver exists, such as constraint programming (CP), mixed integer programming (MIP), or Boolean satisfiability (SAT). In this paper, we enhance existing SAT-based algorithm for MRPP via spartification of the set of candidate paths for each robot from which target Boolean encoding is derived. Suggested sparsification of the set of paths led to smaller target Boolean formulae that can be constructed and solved faster while optimality guarantees of the approach have been kept.

\end{abstract}

\section{Introduction and Motivation}

Multi-robot path planning in graphs (MRPP) represents fundamental problem in combinatorial motion planning in robotics \cite{DBLP:conf/ijcai/LunaB11,DBLP:conf/ijcai/Ryan07,DBLP:conf/aiide/Silver05,DBLP:conf/aaai/Standley10,DBLP:conf/icra/YuL13}. The task is to navigate each robot in the set of robots $R=\{r_1,r_2,...,r_k\}$ from its initial position to a specified goal position. The environment is modeled as an undirected graph $G=(V,E)$ where vertices represent positions and robots move across edges between vertices. Two requirements make the problem challenging: {\bf (1)} the robots must not collide with each other, that is they never can share a vertex or traverse an edge in opposite directions and {\bf (2)} some objective such as total number of actions must be optimized.

Many practical problems from robotics can be interpreted as MRPP. Examples include discrete multi-robot navigation and coordination \cite{DBLP:conf/iros/LunaB10}, robot rearrangement in automated warehouses \cite{DBLP:journals/tase/BasileCC12}, ship collision avoidance \cite{DBLP:journals/jaciii/KimHP14}, or formation maintenance and maneuvering of aerial vehicles \cite{DBLP:conf/icra/ZhouS15}.

We address the MRPP problem from the perspective of compilation-based techniques. {\em Compilation} is one of the most important techniques used across many computing fields ranging from theory to practice. In the context of problem solving in robotics, the compilation approach reduces an input problem instance from its source formalism to a different, usually well established, target formalism for which an efficient solver exists. After obtaining a solution by the solver, it is interpreted back to the input formalism, which altogether constitutes a {\bf reduction-solving-interpretation} loop.

Target formalisms are often combinatorial optimization frameworks like {\em constraint programming / optimization} (CP) \cite{DBLP:books/daglib/0016622}, {\em mixed integer programming} (MIP) \cite{DBLP:books/daglib/0023873,rader2010deterministic}, Boolean satisfiability (SAT) \cite{DBLP:series/faia/2009-185}, satisfiability modulo theories (SMT) \cite{DBLP:reference/mc/BarrettT18}, or answer set programming (ASP) \cite{DBLP:books/sp/Lifschitz19}. Employing the advancements in solvers for the target formalisms, often accumulated for decades, in solving the input problem represents the key benefit of problem solving via compilation. However, the way how the input instance is reduced to the target formalism and presented to the solver has a great impact on the efficiency of the whole solving process.

Currently {\bf compilation-based} optimal solvers for MRPP represent a major alternative to {\bf search-based} solvers, that model and solve the problem directly, and often provide more modular and versatile architecture than search-based solvers while providing competitive performance. Contemporary compilation-based solvers for MRPP include those based on CP \cite{DBLP:conf/aips/GangeHS19}, MIP \cite{DBLP:conf/ijcai/LamBHS19}, SAT/SMT \cite{DBLP:conf/ijcai/Surynek19,DBLP:conf/socs/SurynekFSB16}, as well as ASP-based solvers \cite{DBLP:conf/aaai/ErdemKOS13}.

We focus in this paper on SMT-based solvers for MRPP. Our contribution consists in a new technique for encoding the MRPP instance as Boolean formulae via {\bf sparsification} of a set of candidate paths for each robot. The novel encoding is integrated into a modified SMT-CBS \cite{DBLP:conf/ijcai/Surynek19}, an optimal MRPP solving algorithm that uses lazy compilation scheme. Since Boolean formulae, to which the input MRPP instance is reduced in SMT-CBS, are derived from the set of candidate paths, the effect of sparsification of the set is twofold: {\bf (1)} it leads to smaller target Boolean formulae that can be constructed faster and {\bf (2)} the satisfiability of formulae can be decided by the SAT solver faster, altogether improving the reduction-solving-interpretation loop in SMT-CBS. At the same time, optimality guarantees in the modified SMT-CBS are kept via maintaining easily verifiable property of the sparse set of candidate paths.

The paper is organized as follows: the MRPP problem and basic compilation-based methods for MRPP are introduced first. Then the concept of lazy compilation schemes is described from which the sparsification technique is derived, the main contribution. The new algorithm based on the sparsification technique, called Sparse-SMT-CBS, is developed next and its experimental evaluation follows.

\section{Background}

In this section we introduce the problem of path planning for multiple robots formally and present optimal solving algorithms with focus on compilation-based techniques.

\subsection{Multi-robot Path Planning}

In MRPP, the time is discretized into time steps. The configuration of robots at timestep $t$ is denoted as $s_t$. Each robot $r_i$ has a start position $s_0(r_i) \in V$ and a goal position $s_+(r_i) \in V$.  At each time step a robot can either {\em move} to an adjacent location or {\em wait} in its current location.

Formally, an MRPP instance is a tuple $\Sigma=(G=(V,E),R,s_0,s_+)$ where $s_0:R \rightarrow V$ is an initial configuration of robots and $s_+:R \rightarrow V$ is a goal configuration of robots. A \textit{solution} for $\Sigma$ is a sequence of configurations $\mathcal{S}(\Sigma)=[s_0,s_1,...,s_{\mu}]$ such that $s_{t+1}$ results from valid movements from $s_{t}$ for $t=1,2,...,\mu-1$, and $s_{\mu}=s_+$. Orthogonally, the solution can be represented as a set of paths for individual robots. 

\begin{figure}[h]
    \centering
    \includegraphics[trim={3.2cm 22cm 4.8cm 2.3cm},clip,width=0.80\textwidth]{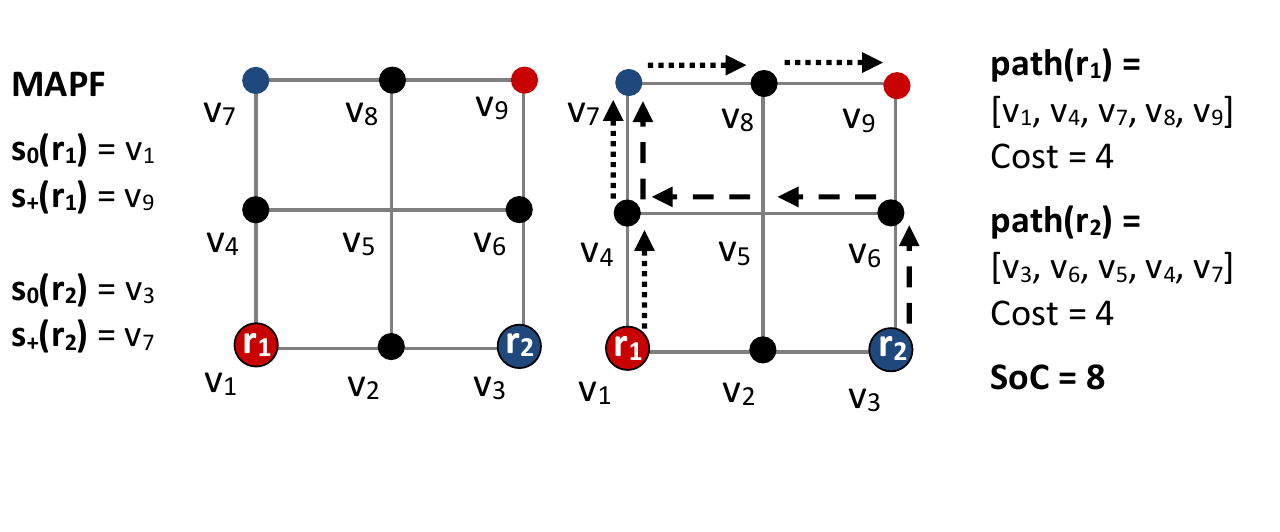}
    \vspace{-1.0cm}\caption{An MRPP instance with two robots $r_1$ and $r_2$.}
    \label{figure-MRPP}
\end{figure}

At each time step a robot can either {\em move} to an adjacent location or {\em wait} in its current location. The task is to find a sequence of move/wait actions for each robot $r_i$, moving it from $s_0(r_i)$ to $s_+(r_i)$ such that robots do not {\em conflict}, i.e., do not occupy the same location at the same time and do not traverse the same edge in opposite directions. An example of MRPP instance and its solution is shown in Figure \ref{figure-MRPP}.

Often various cumulative objectives are optimized in MRPP. We will develop all concepts in this paper for the {\em sum-of-costs} objective, one of the most frequently used, formally defined as follows:

\begin{definition} {\bf (sum-of-costs).}
	{\em Sum-of-costs} denoted $\mathit{SoC}$ is the summation, over all
	robots, of the number of time steps required to reach the goal.
	Formally, $\mathit{SoC} = \sum_{i=1}^k{\mathit{cost}(path(r_i))}$, where $\mathit{cost}(path(r_i))$ is an
	\textit{individual path cost} of robot $r_i$ connecting $s_0(r_i)$ calculated as the number of edge traversals and wait actions. \footnote{The notation $path(r_i)$ refers to path in the form of a sequence of vertices and edges connecting $s_0(r_i)$ and $s_+(r_i)$ while $\mathit{cost}$ assigns the cost to a given path.}
\end{definition}

Observe that in the sum-of-costs we accumulate the cost of wait actions for robots not yet reaching their goal vertices. A feasible solution of a solvable MRPP instance can be found in polynomial time \cite{DBLP:conf/focs/KornhauserMS84}.

Finding an optimal solution with respect to the sum-of-costs objective is NP-hard \cite{DBLP:conf/aaai/Surynek10,DBLP:conf/aaai/YuL13} and also determining the existence of a solution that differs from the optimum by a factor less than $4/3$ is NP-hard too \cite{DBLP:conf/aaai/MaTSKK16}. Therefore designing algorithms based on search and SAT for MRPP is justifiable.

\subsection{Compilation-based Approaches}

The idea behind compilation that uses the SAT paradigm is to construct a Boolean formula whose satisfiability corresponds to existence of a solution of sum-of-costs $\mathit{SoC}$ to a given MRPP $\Sigma$. Moreover, the approach is constructive; that is, the formula exactly reflects the MRPP instance and if satisfiable, solution of MRPP can be reconstructed from satisfying assignment of the formula.

There are two ways how to connect satisfiability of the formula and solvability of $\Sigma$: using either {\bf equivalence} or {\bf implication}.

We say  $\mathcal{F(\mathit{SoC})}$ to be a {\em complete Boolean model} of MRPP.

\begin{definition}
  {\bf (complete Boolean model)} Boolean formula $\mathcal{F(\mathit{SoC})}$ is a {\em complete Boolean model} of MRPP $\Sigma$
  if the following condition holds: $\mathcal{F(\mathit{SoC})}$ is satisfiable $\Leftrightarrow$ $\Sigma$ has a solution of sum-of-costs $\mathit{SoC}$.
\end{definition}

Complete Boolean models were the used in makespan optimal SAT-based solvers for MRPP \cite{DBLP:journals/amai/Surynek17} and in MDD-SAT \cite{DBLP:conf/ecai/SurynekFSB16}, the first sum-of-costs optimal SAT-based solver. A natural relaxation from the complete Boolean model is an {\em incomplete Boolean model} where instead of the equivalence between solving MRPP and the formula we require an implication only. Incomplete models are inspired from the SMT paradigm and are used in the recent sum-of-costs optimal solver SMT-CBS \cite{DBLP:conf/ijcai/Surynek19}.

\begin{definition}
  {\bf (incomplete Boolean model).} Boolean formula $\mathcal{H(\mathit{SoC})}$ is an {\em incomplete Boolean model} of MRPP 
  $\Sigma$ if the following condition holds: $\mathcal{H(\mathit{SoC})}$ is satisfiable $\Leftarrow$ $\Sigma$ has a solution of sum-of-costs $\mathit{SoC}$.
\end{definition}

Being able to construct formula $\mathcal{F}$ one can obtain optimal MRPP solution by checking satisfiability of $\mathcal{F}(0)$, $\mathcal{F}(1)$, $\mathcal{F}(2)$,... until the first satisfiable $\mathcal{F(\mathit{SoC})}$ is met. This is possible due to monotonicity of MRPP solvability with respect to increasing values of common cumulative objectives such as the sum-of-costs. In practice it is however impractical to start at 0; lower bound estimation is used instead - sum of lengths of shortest paths can be used in the case of sum-of-costs.

Construction of $\mathcal{F(\mathit{SoC})}$ relies on the time expansion of underlying graph $G$. Having $\mathit{SoC}$, the basic variant of time expansion determines the maximum number of time steps $\mu$ (also refered to as a {\em makespan}) such that every possible solution of the given MRPP with the sum-of-costs less than or equal to $\mathit{SoC}$ fits within $\mu$ timesteps (that is, no robot is outside its goal vertex after $\mu$-th timestep if the sum-of-costs $\mathit{SoC}$ is not to be exceeded).

The time expansion itself makes copies of vertices $V$ for each timestep $t=0,1,2,...,\mu$. That is, we have vertices $v^t$ for each $v \in V$ time step $t$. Edges from $G$ are converted to directed edges interconnecting timesteps in time expansion. Directed edges $(u^t,v^{t+1})$ are introduced for $t=1,2,...,\mu-1$ whenever there is $\{u,v\} \in E$. Wait actions are modeled by introducing edges $(u^t,t^{t+1})$. A directed path in time expansion corresponds to trajectory of a robot in time. Hence the modeling task now consists in construction of a formula in which satisfying assignments correspond to directed paths from $s_0^0(r_i)$ to $s_+^\mu(r_i)$ in the time expansion.

Assume that we have time expansion $(V_i,E_i)$ for robot $r_i$. Propositional variable $\mathcal{X}_v^t(r_j)$ is introduced for every vertex $v^t$ in $V_i$. The semantics of $\mathcal{X}_v^t(r_i)$ is that it is $\mathit{TRUE}$ if and only if robot $r_i$ resides in $v$ at time step $t$. Similarly we introduce $\mathcal{E}_{u,v}^t(r_i)$ for every directed edge $(u^t,v^{t+1})$ in $E_i$. Analogically the meaning of $\mathcal{E}_{u,v}^t(r_i)$ is that it is $\mathit{TRUE}$ if and only if robot $r_i$ traverses edge $\{u,v\}$ between time steps $t$ and $t+1$.

Finally constraints are added so that truth assignment are restricted to those that correspond to valid solutions of a given MRPP. Added constraints together ensure that $\mathcal{F(\mathit{SoC})}$ is a {\em complete propositional model} for given MRPP.

We here illustrate the model by showing few representative constraints. For the detailed list of constraints we refer the reader to \cite{DBLP:conf/ecai/SurynekFSB16}.

Collisions between robots can be eliminated by the following constraint over $\mathcal{X}_v^t(a_i)$ variables for every $v \in V$ and timestep $t$:

\begin{equation}
    {\sum_{a_i \in A \:|\:v^t \in V_i}{\mathcal{X}^t_v(a_i)} \leq 1
    }
    \label{eq-1}
\end{equation}

Next, there is a constraint stating that if robot $a_i$ appears in vertex $u$ at time step $t$ then it has to leave through exactly one edge $(u^t,v^{t+1})$. This can be established by following constraints:

\begin{equation}
   {  \mathcal{X}_u^t(a_i) \Rightarrow \bigvee_{(u^t,v^{t+1}) \in E_i}{\mathcal{E}^t_{u,v}(a_i),}
   }
   \label{eq:basic-start}
\end{equation}
\begin{equation}
   {  \sum_{v^{t+1}\:|\:(u^t,v^{t+1}) \in E_i }{\mathcal{E}_{u,v}^t{(a_i)} \leq 1}
   }
   \label{eq-2}
\end{equation}

Other constraints ensure that truth assignments to variables per individual robots form paths. That is if robot $r_i$ enters an edge it must leave the edge at the next time step.

\begin{equation}
   {  \mathcal{E}^t_{u,v}(a_i) \Rightarrow \mathcal{X}^t_v(a_i) \wedge \mathcal{X}^{t+1}_v(a_i)
   }
   \label{eq-4}
\end{equation}

A common measure how to reduce the number of decision variables derived from the time expansion is the use of {\em multi-valued decision diagrams} (MDDs) \cite{DBLP:conf/cp/AndersenHHT07,DBLP:journals/ai/SharonSGF13}. The basic observation that holds for MRPP is that an robot can reach vertices in the distance $d$ (distance of a vertex is measured as the length of the shortest path) from the current position of the robot no earlier than in the $d$-th time step. Analogical observation can be made with respect to the distance from the goal position.

\begin{figure}[h]
    \centering
    \vspace{-0.2cm}
    \includegraphics[trim={2.7cm 16.2cm 2.8cm 3cm},clip,width=0.80\textwidth]{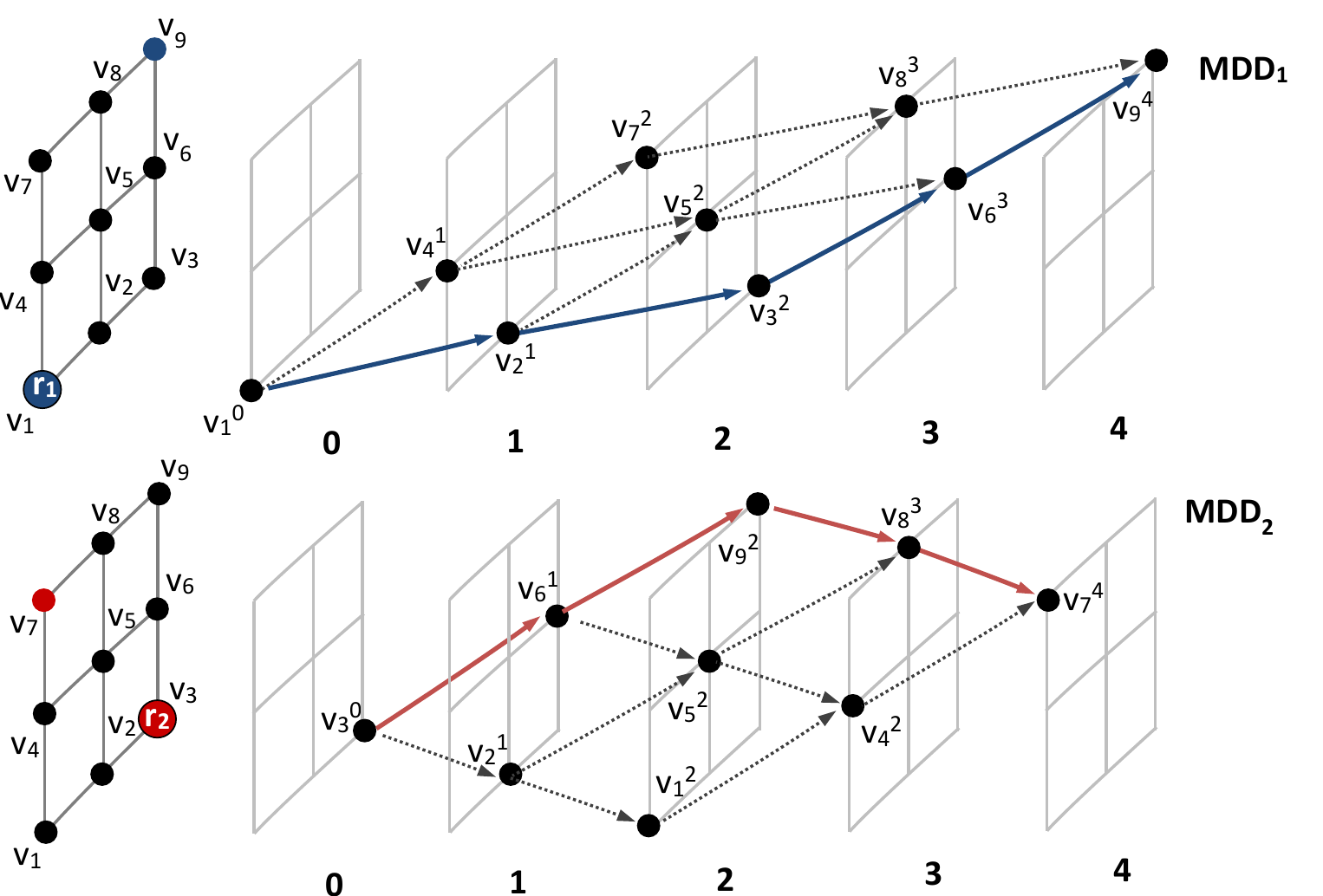}
    \vspace{-0.2cm}
    \caption{An example of MDDs representing all candidate paths of cost 4 for robots $r_1$ nad $r_2$.}
    \label{fig-MDD}
\end{figure}

Above observations can be utilized when making the time expansion of $G$. For a given robot, we do not need to consider all vertices at time step $t$ but only those that are reachable in $t$  timesteps from the initial position and that ensure that the goal can be reached in the remaining $\mu - t$ timesteps. Time expansion using MDDs is shown in Figure \ref{fig-MDD}.

The combination of SAT-based approach and MDD time expansion led to the MDD-SAT algorithm.

\subsection{CBS and Lazy Compilation}

{\em Conflict-based search} (CBS) \cite{DBLP:journals/ai/SharonSFS15} is a popular algorithm for solving MRPP optimally. From the compilation perspective, the CBS algorithm could be understood as a {\bf  lazy} method that tries to solve an underspecified problem and relies on to be lucky to find a correct solution even using this incomplete specification. There is another mechanism that ensures soundness of this lazy approach, the branching scheme. If the CBS algorithm is not lucky, that is, the candidate solution is incorrect in terms of MRPP rules, then the search branches for each possible refinement of discovered MRPP rule violation and the refinement is added to the problem specification in each branch. Concretely, the MRPP rule violations are conflicts of pairs of robots such as collision of $a_i \in A$ and $a_j \in A$ in $v$ at time step $t$ and the refinements are conflict avoidance constraints for single robots in the form that $a_i \in A$ should avoid $v$ at time step $t$ (for $a_j$ analogously).

Analogously the avoidance for edge conflicts can be done. For the sake of brevity, all the concepts developed further will be introduced only for vertex conflicts.

\begin{algorithm}[h]
\begin{footnotesize}
\SetKwBlock{NRICL}{SMT-CBS ($\Sigma = (G=(V,E),R,s_0,s_+))$}{end} \NRICL{
    $\mathit{conflicts} \gets \emptyset$\\
    $\mathit{paths} \gets$ $\{\mathit{path}^*(r_i)$ a shortest path from $s_0(r_i)$ to $s_+(r_i) | i = 1,2,...,k\}$ \\
    $\mathit{SoC} \gets \sum_{i=1}^k{\mathit{cost}(\mathit{paths}(r_i))}$ \\
    $\mu \gets \max_{i=1}^k{\mathit{cost}(\mathit{paths}(r_i))}$ \\    
    \While {$\mathit{TRUE}$}{
        $(\mathit{paths},\mathit{conflicts}) \gets$ SMT-CBS-Fixed($\mathit{conflicts},\mu,\mathit{SoC},\Sigma$)\\
        \If {$\mathit{paths} \neq \mathit{UNSAT}$}{
        	\Return $\mathit{paths}$\\
        }
        $\mathit{SoC} \gets \mathit{SoC} + 1$ \\
        $\mu \gets \mu + 1$\\
    }
}   
 
\SetKwBlock{NRICL}{SMT-CBS-Fixed($\mathit{conflicts},\mu,\mathit{SoC},\Sigma$)}{end} \NRICL{
          $\mathcal{F}(\mathit{SoC}) \gets$ encode-Incomplete$(\mathit{conflicts},\mu,\mathit{SoC},\Sigma)$\\
	    \While {$\mathit{TRUE}$}{
	        $\mathit{assignment} \gets$ consult-SAT-Solver$(\mathcal{F}(\mathit{SoC}))$\\
	        
	        \If {$\mathit{assignment} \neq UNSAT$}{
	            $\mathit{paths} \gets$ extract-Solution$(\mathit{assignment})$\\
	            $\mathit{collisions} \gets$ validate($\mathit{paths}$)\\
                   \If {$\mathit{collisions} = \emptyset$}{
                      \Return $(\mathit{paths},\mathit{conflicts})$\\
                   }
                   \For{each $(r_i,r_j,v,t) \in \mathit{collisions}$}{
                      $\mathcal{F}(\mathit{SoC}) \gets \mathcal{F}(\mathit{SoC}) \cup \{\neg \mathcal{X}_v^t(r_i) \vee \neg \mathcal{X}_v^t(r_j)$\}\\
                      $\mathit{conflicts} \gets \mathit{conflicts} \cup \{[(r_i,v,t),(r_j,v,t)]\}$
                   }
               }
               \Else{
                   \Return {($\mathit{UNSAT},\mathit{conflicts}$)}\\
               }
          }
}
\caption{SMT-based optimal MRPP solver} \label{alg-SMTCBS}
\end{footnotesize}
\end{algorithm}

The idea of laziness of CBS has been combined with compilation-based approaches in SMT-CBS algorithm presented using pseudo-code as Algorithm \ref{alg-SMTCBS}. The algorithm is divided into two procedures: SMT-CBS representing the main loop and SMT-CBS-Fixed solving the input MRPP for a fixed cost $\mathit{SoC}$. The major difference from the standard CBS is that there is no branching at the high level. The high level SMT-CBS roughly correspond to the main loop of MDD-SAT. The set of conflicts is iteratively collected during the entire execution of the algorithm. The incomplete Boolean model is constructed from MDDs by {\em encode-Incomplete} that produces encoding from MDD-SAT that ignores specific collisions between robots.

The conflict resolution in standard CBS implemented as high-level branching is here represented by refinement of $\mathcal{F}(\mathit{SoC})$ with disjunction (line 22). Branching is thus deferred into the SAT solver. The advantage of SMT-CBS is that it builds the formula lazily; that is, it adds constraints on demand after a conflict occurs. Such approach may save resources as solution may be found before all constraint are added.


\section{Sparsification in Lazy Compilation}

Although lazy compilation in SMT-CBS is the major factor that reduces the size of generated Boolean formulae during the solving process compared to MDD-SAT, the reduction concerns only the conflict avoidance constraints for pairs of robots (binary disjunctions). The number of nodes as well as the number of directed edges in the MDDs is not reduced by laziness and hence the number of variables and constraints modeling the existence of directed paths in MDDs in the target Boolean formula is the same as in MDD-SAT. For instances that take place on large graphs, the size of MDDs directly reflected in the size of the formula could still be prohibitive.

Therefore we suggest to simplify the formulae by reducing the number of directed paths that are encoded. To do this, we suggest to use sparse sets of candidate paths for each robot that satisfy given sum-of-costs and makespan bounds. That is, instead of considering all such paths as done in MDDs we consider only a relevant subset of them. The sparse set of candidate paths will be denoted $\Pi$. Similar concept called the {\em pool of paths} has been used in the context of MIP-based compilation for MRPP \cite{DBLP:conf/aips/GangeHS19} which however does not explicitly focus on sparsification.

We integrated the sparse paths set reasoning into the SMT-CBS framework, designing a new algorithm we called {\bf Sparse-SMT-CBS}. The pseudo-code of Sparse-SMT-CBS is shown as Algorithm \ref{alg-sparse-SMTCBS}.

Sparse-SMT-CBS uses identical sum-of-costs and makespan bounds increasing scheme as SMT-CBS at the high-level (lines 1-11). Each iteration at the high-level resolves a question whether there exists a solution to the input MRPP $\Sigma$ such that it fits in the current sum-of-costs $\mathit{SoC}$ and makespan $\mu$. This question is compiled as a series of Boolean formulae and consulted with the SAT solver at the lower-level (lines 12-30).

The lower-level in Sparse-SMT-CBS is different from SMT-CBS. In both algorithms it tries to find a non-conflicting set of paths satisfying $\mathit{SoC}$ and $\mu$, but the set of candidate paths from which SMT-CBS selects is fixed in advance in MDD, while Spare-SMT-CBS starts with a minimal set of candidate paths and each time a new conflict is discovered the set of candidate paths is extended to reflect the new conflict. To keep soundness and optimality of the algorithm, the sparse set of candidate paths must be selected to according to the following conditions.

\begin{definition} {\bf (path feasability).}
Let $C$ be a set of conflicts of the form $\{r_i,v,t\}$ forbidding robot $r_i \in R$ to reside in $v \in V$ at timestep $t$. We say a path $\mathit{path}(r_i) = [p_0, p_1, ..., p_m]$ for robot $r_i$ to be {\em feasible} with respect to $C$ if and only if $\{r_i, p_t, t\} \notin C$ for $\forall t \in \{0,1,...,m\}$.
\end{definition}

The sparse set of paths for robot $r_i$ denoted $\Pi(r_i)$ with respect to a set of conflicts $C$ and makespan $\mu$ must satisfy the following property:

\begin{itemize}
\item {\bf (P1)} $\Pi(r_i)$ contains at least one feasible path with respect to each subset of conflicts $C' \subseteq C$.
\end{itemize}

The property (P1) must be reflected in implementation of new-Paths function called at line 25. Intuitively said, (P1) ensures that after discovering a new conflict at least one path avoiding the new conflict for each possible combination of previous conflicts is added to $\Pi$ provided that such path exists for given makespan bound $\mu$.

\begin{algorithm}[h]
\begin{footnotesize}
\SetKwBlock{NRICL}{Sparse-SMT-CBS ($\Sigma = (G=(V,E),R,s_0,s_+))$}{end} \NRICL{
    $\mathit{conflicts} \gets \emptyset$\\
    $\mathit{\Pi} \gets$ $\{\mathit{\Pi}^*(r_i)$ a shortest path from $s_0(r_i)$ to $s_+(r_i) | i = 1,2,...,k\}$ \\
    $\mathit{SoC} \gets \sum_{i=1}^k{\mathit{cost}(\mathit{\Pi}(r_i))}$ \\
    $\mu \gets \max_{i=1}^k{\mathit{cost}(\mathit{\Pi}(r_i))}$ \\    
    \While {$\mathit{TRUE}$}{
         $(\mathit{path},\mathit{conflicts}) \gets$ Sparse-SMT-CBS-Fixed($\mathit{conflicts},\mathit{\Pi},\mu,\mathit{SoC},\Sigma$)\\
        \If {$\mathit{paths} \neq \mathit{UNSAT}$}{
        	\Return $\mathit{paths}$\\
        }
        $\mathit{SoC} \gets \mathit{SoC} + 1$ \\
        $\mu \gets \mu + 1$\\        
    }
}   
 
\SetKwBlock{NRICL}{Sparse-SMT-CBS-Fixed($\mathit{\Pi},\mathit{conflicts},\mu,\mathit{SoC},\Sigma$)}{end} \NRICL{
	    \While {$\mathit{TRUE}$}{
	        $\mathcal{F}(\mathit{SoC}) \gets$ encode-Incomplete$(\mathit{\Pi}, \mathit{conflicts},\mu,\mathit{SoC},\Sigma)$\\	    
	        $\mathit{assignment} \gets$ consult-SAT-Solver$(\mathcal{F}(\mathit{SoC}))$\\
	        \If {$\mathit{assignment} \neq \mathit{UNSAT}$}{
	            $\mathit{paths} \gets$ extract-Solution$(\mathit{assignment})$\\
	            $\mathit{collisions} \gets$ validate($\mathit{paths}$)\\
                   \If {$\mathit{collisions} = \emptyset$}{
                      \Return $(\mathit{paths},\mathit{conflicts})$\\
                   }
                   \For{each $(r_i,r_j,v,t) \in \mathit{collisions}$}{
                      $\mathcal{F}(\mathit{SoC}) \gets \mathcal{F}(\mathit{SoC}) \cup \{\neg \mathcal{X}_v^t(r_i) \vee \neg \mathcal{X}_v^t(r_j)$\}\\
                      $\mathit{conflicts} \gets \mathit{conflicts} \cup \{[(r_i,v,t),(r_j,v,t)]\}$
                   }
                   \For{each $r_i \in R$}{
                      $\pi(r_i) \gets$ new-Paths($\mathit{\Pi}(r_i), \mathit{conflicts}(r_i),\mu$) \\
                      $\mathit{\Pi}(r_i) \gets \mathit{\Pi}(r_i) \cup \pi(r_i)$
                   }
                   \If {$\bigcup_{i=1}^k{\pi(r_i)} = \emptyset$}{
                   		\Return {($\mathit{UNSAT}, \mathit{conflicts}$)} \\
                   }
               }
               \Else{
                   \Return {($\mathit{UNSAT}, \mathit{conflicts}$)}\\
               }
          }
}
\caption{SMT-based optimal MRPP solver with sparsification of the set of candidate paths.} \label{alg-sparse-SMTCBS}
\end{footnotesize}
\end{algorithm}

The sparse set of candidate paths is used to build an incomplete Boolean MRPP model $\mathcal{F}(\mathit{SoC})$ (line 14) which is subsequently solved by the SAT solver. If there is no satisfying truth-value assignment for $\mathcal{F}(\mathit{SoC})$ (line 30), then we can conclude taking into account (P1) that there is no solution of the input MRPP satisfying the $\mathit{SoC}$ and $\mu$ bounds and the control returns to the high-level where the bounds are incremented (lines 10-11). If a satisfying truth-value assignment of $\mathcal{F}(\mathit{SoC})$ is found (line 16), then it has to be interpreted and checked against MRPP rules, that is, it is checked for collisions between robots (line 18). If there are no collisions, the algorithm can return a valid MRPP solution (line 20). If a collision is detected (lines 21-28), a proper treatment must be applied.

This includes {\bf (1)} extending the model $\mathcal{F}(\mathit{SoC})$ with a collision elimination constraint, a disjunction forbidding a pair of conflicting actions to take place at once (lines 21-23) and {\bf (2)} (new to Sparse-SMT-CBS) and extending the sparse set of candidate paths $\Pi$ with new paths reflecting newly discovered conflicts (line 24-26). It may however happen that no new path could be found any robot which again together with property (P1) means that there is no solution for current $\mathit{SoC}$ and $\mu$ bounds (lines 27-28) and control returns to the high-level to continues with the next iteration.

Without proof let us state the following proposition summarizing the soundness and optimality of the Sparse-SMT-CBS algorithm. Let us note that unsolvable input MRPP can be detected in advance by some polynomial-algorithm such as Push-and-Rotate \cite{DBLP:journals/jair/WildeMW14}.
\vspace{0.2cm}
\begin{proposition}
Sparse-SMT-CBS finds sum-of-costs optimal solution for a solvable input MRPP $\Sigma$ provided the sparse set of candidate paths $\Pi$ satisfies property (P1). $\blacksquare$
\end{proposition}
\vspace{0.2cm}

Paths included into $\Pi$ can be found by various shortest path algorithms like A* that take into account the set of conflicts. Moreover deciding what paths will be preferably included in $\Pi$ provides a room to integrate MRPP specific heuristics such as {\em cardinal conflict} aggregation \cite{DBLP:conf/ijcai/LiFB0K19} or {\em symmetry breaking} \cite{DBLP:conf/aips/0001GHS0K20} which is known to be difficult for MDD-SAT and SMT-CBS.

Notice also that property (P1) is used is an argument for correct functionality of the algorithm only when it detects non-existence of solution for the current bounds. Except these two moments the property can be relaxed which could provide greater room for integrating MRPP specific heuristics.

\subsection{Details of the Boolean Model}

Eventually it may happen that the sparse set of paths grows during the course of low-level phase so that all possible paths satisfying given sum-of-costs and makespan bounds are eventually included. In such a case, the compression based on MDDs is useful similarly as in SMT-CBS or MDD-SAT. The concept of MDD is useful even for representing sparse sets of paths since often to avoid a new conflict, a new path arriving to the set differs only little from already represented paths.

Therefore we extend the usage of MDDs on sparse sets of candidate paths. To distinguish MDDs representing sparse sets we denote them as {\em sparse MDDs} (SMDDs). Given maximum number of steps $\mu$ and any (sparse) set of paths $\Pi(r_i)$ in the time expansion of $G$ of length at most $\mu$, we are able to construct SMDD$_i$ that represents $\Pi(r_i)$. That is, any path from $\Pi(r_i)$ is represented in SMDD$_i$ and every path represented in SMDD$_i$ is an element of $\Pi(r_i)$.

The encoding used in Sparse-SMT-CBS follows MDD-SAT with collision avoidance constraints omitted but instead of deriving decision Boolean variables and constraints from MDDs they are derived from SMDDs. That is, a Boolean variable $\mathcal{X}(r_i)_v^t$ is introduced for each node $v^t$ in SMDD$_i$ and a variable $\mathcal{E}(r_i)_{u,v}^t$ is introduced for each edge in SMDD$_i$. Constraints are introduced analogously. Observe that the sparser the SMDDs are the smaller Boolean model is obtained.

\begin{figure}[h]
    \centering
    \vspace{-0.2cm}
    \includegraphics[trim={2.5cm 16.4cm 1.6cm 2.6cm},clip,width=0.90\textwidth]{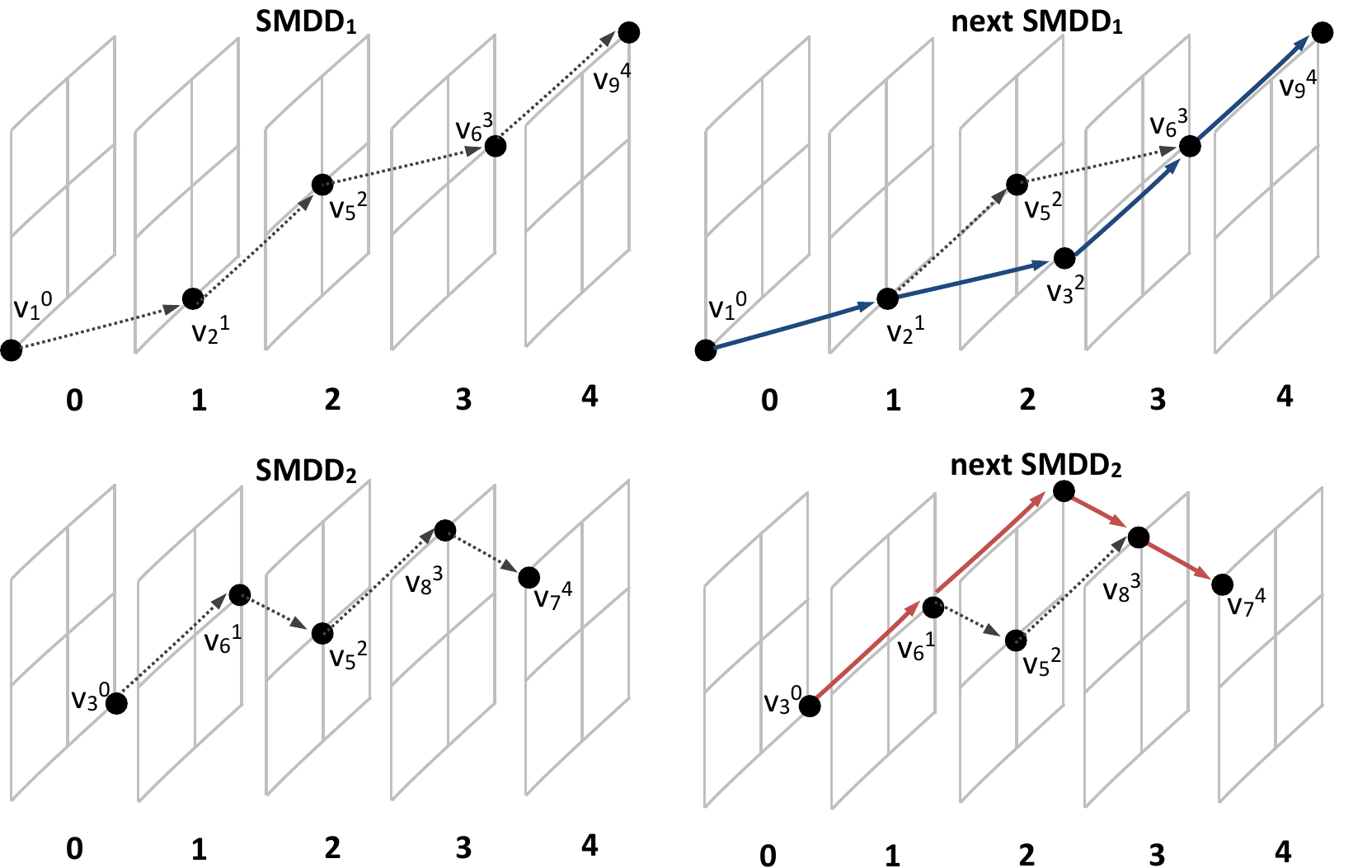}
    \vspace{-0.2cm}
    \caption{An example of sparse MDDs (SMDDs) for robots $r_1$ and $r_2$. The first iteration yields a conflict in $v_5$ at timestep 2 between the robots which can be avoided via newly represented paths in the next iteration.}
    \label{fig-sparseMDD}
\end{figure}

SMDDs from the running example are shown in Figure \ref{fig-sparseMDD}. Initially shortest paths are included in $\Pi(r_i)$ connecting robot's starting vertex and its goal, one shortest path per robot. However, the initial choice of paths is poor leading to a collision in $v_5$ at timestep 2 which is detected by Sparse-SMT-CBS and alternative paths reflecting the conflict are suggested for each robot by the new-Paths function. The new paths are included in SMDDs as shown in the right part of Figure \ref{fig-sparseMDD}. For these new SMDDs, non-conflicting path can be found.

\section{Experimental Evaluation}

We performed an extensive evaluation of the SMT-CBS and Sparse-SMT-CBS algorithms on standard benchmarks from \texttt{movingai.com} \cite{DBLP:conf/ijcai/BoyarskiFSSTBS15,DBLP:journals/ai/SharonSGF13,DBLP:journals/tciaig/Sturtevant12}. To provide broader perspective we also include comparison with the CBS algorithm, a search-based algorithm. Representative part of results is presented in this section.

\subsection{Benchmarks and Setup}

We implemented Sparse-SMT-CBS in C++ using the existing implementation of SMT-CBS. Both algorithms are built on top of the Glucose 3.0 SAT solver \cite{DBLP:journals/ijait/AudemardS18} that still ranks among the best SAT solvers according to recent SAT solver competitions \cite{DBLP:conf/aaai/BalyoHJ17}. The search for new candidate paths that reflect new conflict is done via A* algorithm that takes into account the current set of conflicts. Each conflict is both avoided and ignored during the search by A* resulting in finding paths with respect to all subsets of conflicts. Moreover the strategy of path generation is that priority is given to path reflecting fewer conflicts, starting with path for single conflict, while paths reflecting two and more conflicts are considered later. This turned out to be a good heuristic according to preliminary experiments.

As for CBS we used implementation from \cite{DBLP:conf/socs/BarrerSSF14} written in C\#. Although this implementation of CBS does not use recent MRPP heuristics like rectangle reasoning \cite{DBLP:conf/ijcai/LiFB0K19} we consider it suitable for our experiments as implementations of both Sparse-SMT-CBS and SMT-CBS also do not use any MRPP specific heuristics. Altogether all implementations in the conducted evaluation are based only on the core idea of respective algorithm.

The SAT solver we used supports incremental mode which means that when the formula is extended with new variables and clauses the solver does not need to start from scratch but it can use learned clauses from its previous runs. This incremental mode is highly utilized across the low-level phase of both SMT-CBS and Sparse-SMT-CBS where the target Boolean formula is extended. However, each new iteration of the main loop starts with new instance of the SAT solver since establishing aleviated cost bounds cannot be easily done incrementally.

All experiments were run on system consisting of Xeon 2.8 GHz cores, 32 GB RAM, running Ubuntu Linux 18. \footnote{To enable reproducibility of presented results we will provide complete source code of our solvers and detailed experimental data on author's web: \texttt{http://users.fit.cvut.cz/$\sim$surynpav/iros2021}.}

The experimental evaluation has been done on diverse instances consisting of 4-connected {\em grid} maps ranging in sizes from small to relatively large.

Grid maps are used for discretization of real environments. Robots traverse map using orthogonal movements (diagonal moves are not used) hence unit time per move can be realistically assumed. Moreover this discretization turned out to be sufficient for execution of plans with real robots as shown in our related research on simulations with small mobile robots \cite{DBLP:conf/smc/ChudyPS20}.

\begin{figure}[t]
    \centering
    \includegraphics[trim={2.5cm 14.5cm 2cm 2.5cm},clip,width=1.0\textwidth]{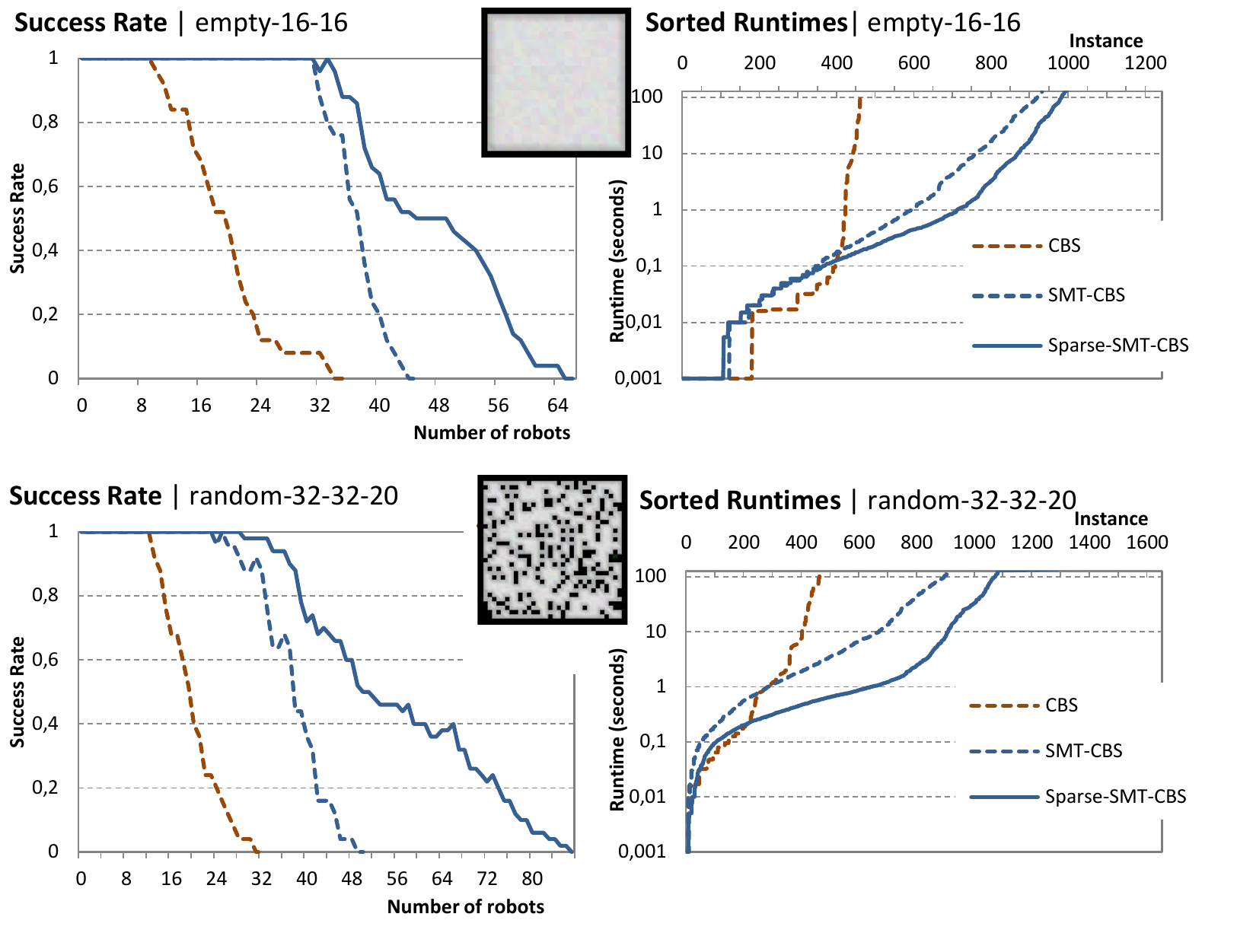}
    \caption{Success rate and runtime comparison on small-sized maps.}
    \label{expr-small}
\end{figure}

We varied the number of robots to obtain instances of various difficulties while initial and goal configurations of robots were generated according to scenarios provided on \texttt{movingai.com}. Depending on the map size, we generated instances with up to 64 (small maps) or 128 (large maps) robots and 25 different instances per number of robots. The time limit in all test was set to 128 seconds. Presented results were obtained from instances solved within this timeout.

\subsection{Runtime Results}

\begin{figure}[h]
    \centering
    \includegraphics[trim={2.5cm 14.5cm 2cm 2.5cm},clip,width=1.0\textwidth]{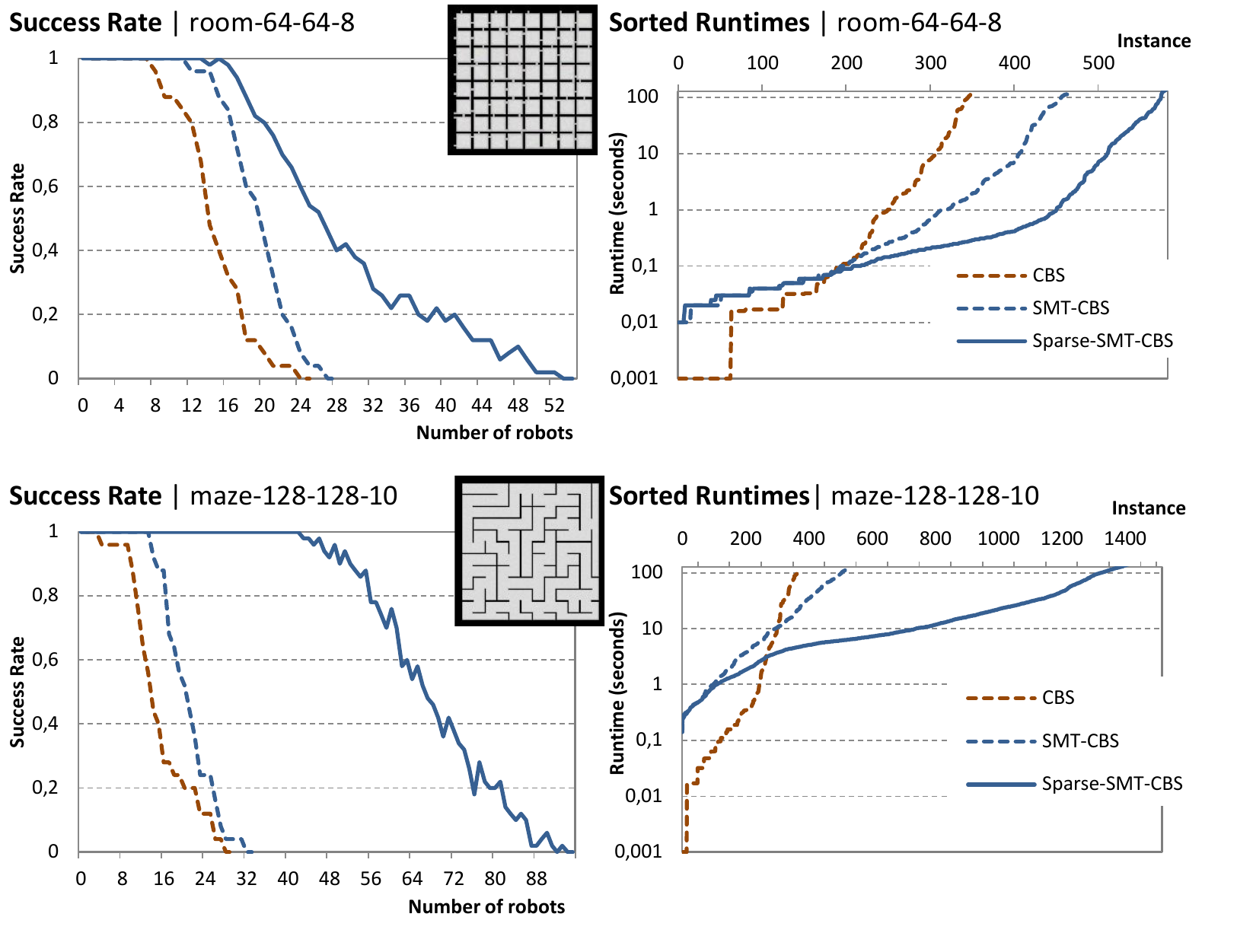}
    \caption{Success rate and runtime comparison on medium-sized maps.}
    \label{expr-medium}
\end{figure}

Results are presented in Figures \ref{expr-small}, \ref{expr-medium}, and \ref{expr-large}. We present {\em success rate} and {\em sorted runtimes}. Success rate shows the ratio of instances solved under the time limit of 128 seconds out of 25 instances per number of robots. Sorted runtimes are inspired by the cactus plots from the SAT Competition \cite{DBLP:conf/aaai/BalyoHJ17}.

The cactus plot for rumtime has been generated by taking runtimes of all instances solved under the time limit by a given algorithm and sorting them along x-axis; so the x-th data-point represents the runtime of x-th easiest instance for the given algorithm. The faster algorithm typically yields to a lower curve in the cactus plot but differences for various regions of the plot can be observed too.

The general trend observable across all maps from success rate and sorted runtimes is that Sparse-SMT-CBS represents a significant improvement over SMT-CBS. The performance gap between SMT-CBS and Sparse-SMT-CBS grows with growing number of robots. The second trend that is observable in the results is that for larger maps (\texttt{maze-128-128-10} and \texttt{lak303d}) the advantage of Sparse-SMT-CBS is greater than in small maps (\texttt{empty-16-16} and \texttt{random-32-32-10}).

Sorted runtimes comparison also shows that in easy instances the sparsification technique does not help significantly. However as the difficulty of instances grows it starts to be increasingly beneficial.

Comparison with CBS shows an expectable trend that due to its negligible overhead, the CBS algorithm dominates in easy instances where both Sparse-SMT-CBS and SMT-CBS need to deal with complex instance compilation process that eventually does not pay off. As difficulty of instances grow, the advanced learning mechanisms and intelligent space search pruning implemented in SAT solvers has greater impact resulting in better performance of SMT-CBS and Sparse-SMT-CBS in more difficult instances.

The known disadvantage of SMT-CBS is that in large maps like {\texttt{warehouse- 10-20-10-2-1} or {\texttt{lak303d} it generate a formulae that are too large which causes degradation in performance. According to our experiments it seems that sparsification in SMT-CBS helps the solver especially on large maps. The reason is that the size of MDDs for large maps is often prohibitive and causes significant difficulty for the SAT solver as it needs to deal with a large formula derived from MDD. Sparsification enables choosing few promising paths that are represented in SMDDs which improves both the Boolean formula construction process as well as its solving.

\begin{figure}[h]
    \centering
    \includegraphics[trim={2.5cm 14.5cm 2cm 2.5cm},clip,width=1.0\textwidth]{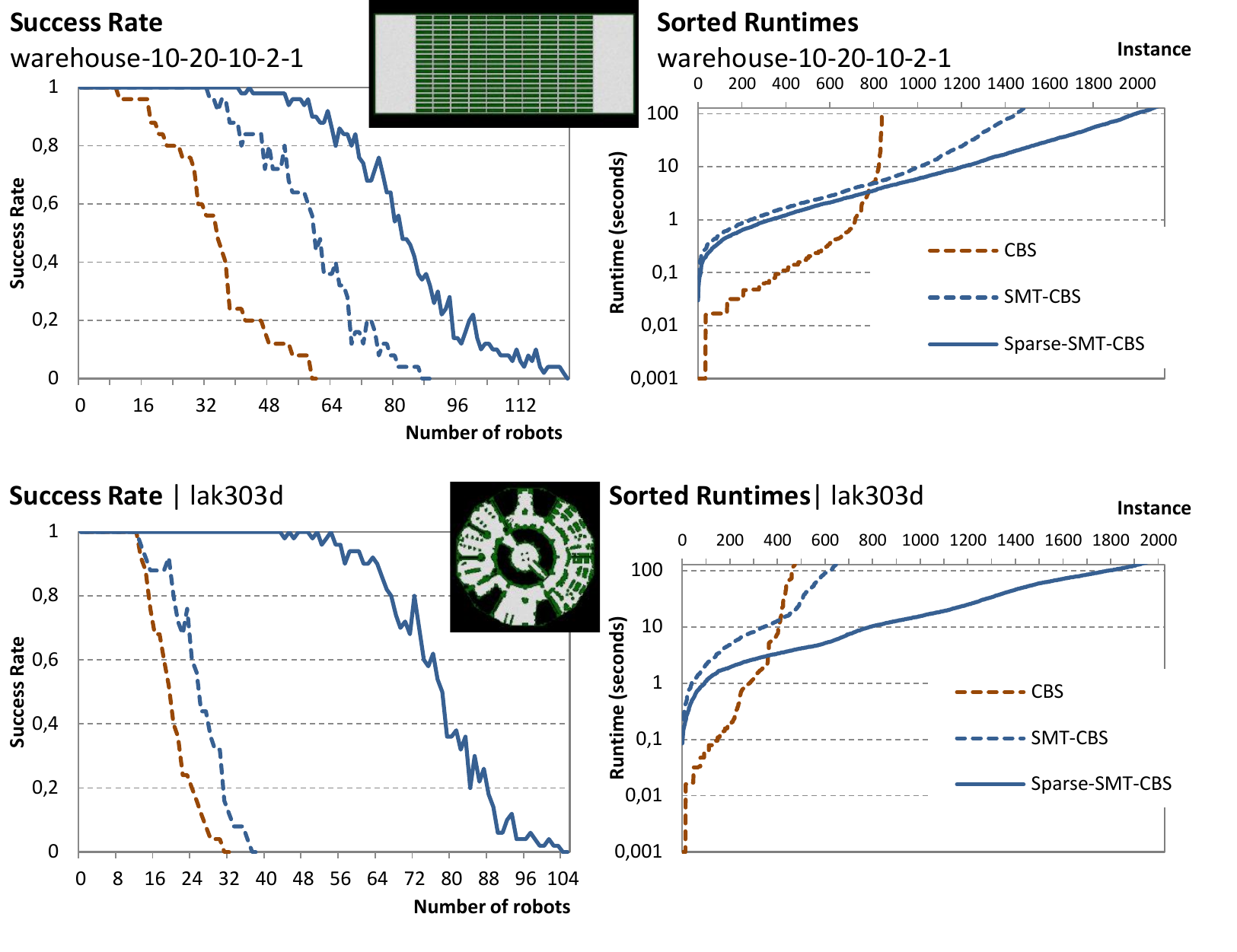}
    \caption{Success rate and runtime comparison on large-sized maps.}
    \label{expr-large}
\end{figure}

\section{Conclusion}

We suggested a technique of sparsification of the set of candidate paths in the context of SAT-based approach to multi-robot path planning (MRPP). The technique aims on currently the most significant drawback of compilation-based approaches to MRPP which is the performance on large graphs. The sparsification technique suggests to construct the target Boolean formula according to a sparse set of candidate paths for each robot which leads to constructing smaller formulae that can be answered by the SAT solver faster. We integrated sparsification into SMT-CBS, a state-of-the-art SAT-based optimal algorithm for MRPP, while keeping its optimality guarantees.

According to our experiments on a number of benchmarks, the new algorithm called Sparse-SMT-CBS performs significantly better than SMT-CBS especially on MRPPs with large graphs and outperforms basic search-based CBS.

Another important advantage of sparsification is that it provides a room for integrating domain specific heuristics via giving a preference to some paths being selected into the set of candidate paths for a robot. Integration of heuristics is difficult in previous SAT-based MRPP solvers like MDD-SAT and SMT-CBS.

For future work we plan to further generalize the concept of sparsification for different variants of MRPP.

\section*{\uppercase{Acknowledgment}}
\noindent
This work has been supported by GA\v{C}R - the Czech Science Foundation, grant registration number 19-17966S.

%
%
%
\bibliographystyle{splncs04}
\bibliography{references}

\end{document}